\documentclass[runningheads]{llncs}
\usepackage{caption}
\usepackage{graphicx}
\usepackage{comment}
\usepackage{times}
\usepackage{epsfig}
\usepackage{diagbox}
\usepackage{mathtools}
\usepackage{enumitem}
\usepackage{hyperref}
\usepackage{amsmath,amssymb} % define this before the line numbering.
\usepackage{color}
\usepackage[linesnumbered,ruled,vlined]{algorithm2e}
\usepackage[font=small,labelfont=bf]{caption}
\usepackage[compact]{titlesec}  
\DeclareMathOperator*{\argmin}{arg\,min}

\begin{document}
% \linenumbers
\pagestyle{headings}
\mainmatter
\def\ECCVSubNumber{5643}  % Insert your submission number here

\title{A Broader Study of \\ Cross-Domain Few-Shot Learning} % Replace with your title

\begin{comment}
\titlerunning{ECCV-20 submission ID \ECCVSubNumber} 
\authorrunning{ECCV-20 submission ID \ECCVSubNumber} 
\author{Anonymous ECCV submission}
\institute{Paper ID \ECCVSubNumber}
\end{comment}
%******************

%\begin{comment}
\titlerunning{A Broader Study of \\ Cross-Domain Few-Shot Learning}
\author{
Yunhui Guo$^{*}$ \inst{1} \
\and 
Noel C. Codella$^{*}$ \inst{2} \
\and
Leonid Karlinsky \inst{2} \ 
\and
James V. Codella \inst{2} \
\and
John R. Smith \inst{2} \ 
\and
Kate Saenko \inst{3} \ 
\and
Tajana Rosing \inst{1} \
\and
Rogerio Feris \inst{2}
}
\authorrunning{Guo, Yunhui, et al.}
\institute{University of California San Diego \and IBM Research AI\and Boston University}
%\end{comment}
%******************
\maketitle

\begin{abstract}
Recent progress on few-shot learning largely relies on annotated data for meta-learning: base classes sampled from the same domain as the novel classes. However, in many applications, collecting data for meta-learning is infeasible or impossible. This leads to the cross-domain few-shot learning problem, where there is a large shift between base and novel class domains. While investigations of the cross-domain few-shot scenario exist, these works are limited to natural images that still contain a high degree of visual similarity. No work yet exists that examines few-shot learning across different imaging methods seen in real world scenarios, such as aerial and medical imaging. In this paper, we propose the Broader Study of Cross-Domain Few-Shot Learning (BSCD-FSL) benchmark, consisting of image data from a diverse assortment of image acquisition methods. This includes natural images, such as crop disease images, but additionally those that present with an increasing dissimilarity to natural images, such as satellite images, dermatology images, and radiology images. Extensive experiments on the proposed benchmark are performed to evaluate state-of-art meta-learning approaches, transfer learning approaches, and newer methods for cross-domain few-shot learning. The results demonstrate that state-of-art meta-learning methods are surprisingly outperformed by earlier meta-learning approaches, and all meta-learning methods underperform in relation to simple fine-tuning by 12.8\% average accuracy. In some cases, meta-learning even underperforms networks with random weights. Performance gains previously observed with methods specialized for cross-domain few-shot learning vanish in this more challenging benchmark. Finally, accuracy of all methods tend to correlate with dataset similarity to natural images, verifying the value of the benchmark to better represent the diversity of data seen in practice and guiding future research. Code for the experiments in this work can be found at  \url{https://github.com/IBM/cdfsl-benchmark}.
%Code for all experiments in this work will be made available on GitHub.
%(temporary anonymous link: https://drive.google.com/file/d/1SJ0lu9uQ6iNy-MG3iXOy\_DdcilLpCZmX/view?usp=sharing).
%In summary, the proposed benchmark serves as a valuable platform to guide future research on cross-domain few-shot learning due to its broad coverage, its challenging nature, and its relation to real-world applications.

%Finally, we demonstrate that transferring from multiple pre-trained models achieves best performance, with accuracy improvements of 14.9\% and 1.9\% versus the best of meta-learning and single model fine-tuning approaches, respectively. 

\end{abstract}
\let\thefootnote\relax\footnote{$^{*}$ Equal contribution.}

%%%%%%%%% BODY TEXT
\section{Introduction}
%\blfootnote{$^*$ Equal contribution}
\begin{figure}[!h]
    \centering
    \includegraphics[width=0.8\linewidth]{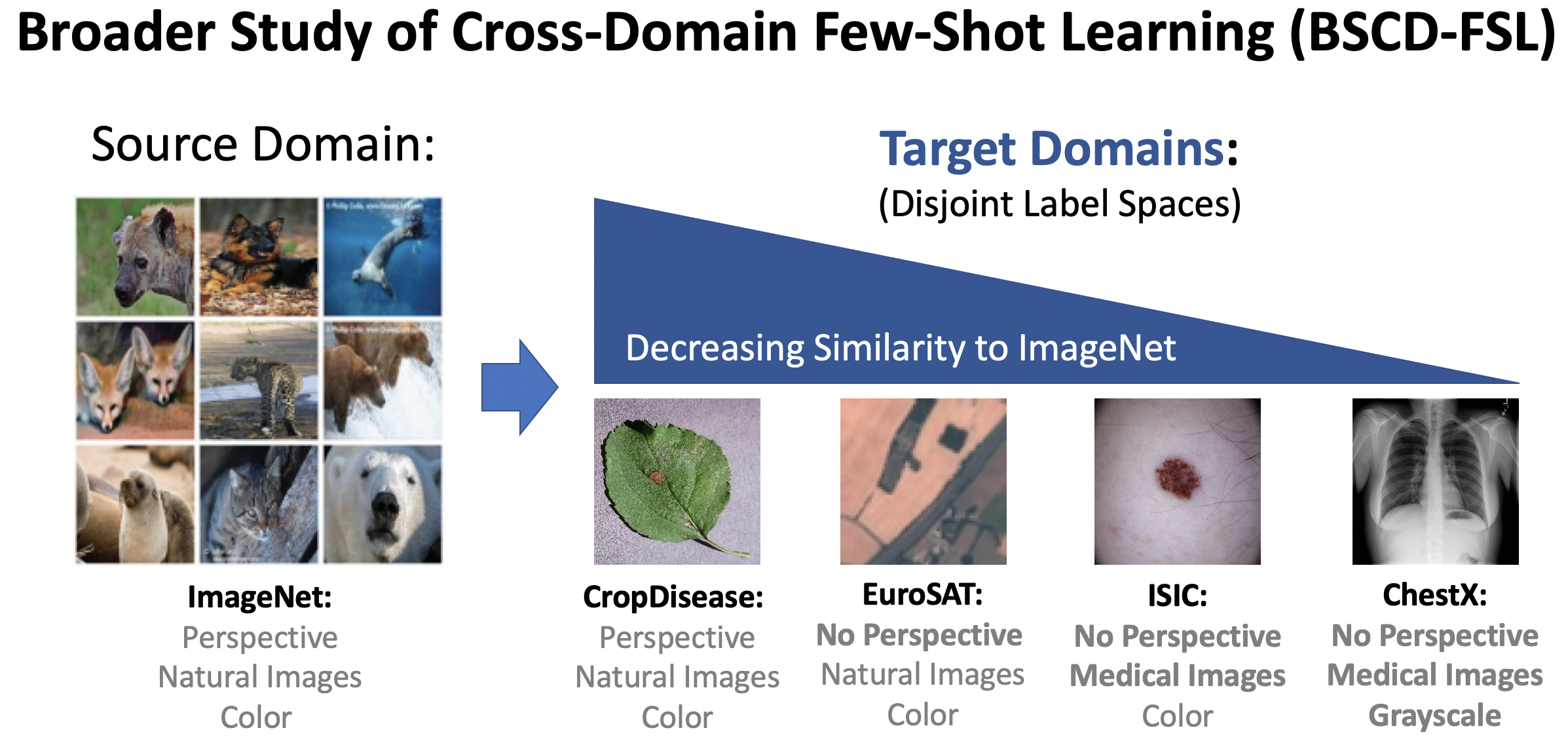}
    \caption{The Broader Study of Cross-Domain Few-Shot Learning (BSCD-FSL) benchmark. ImageNet is used for source training, and domains of varying dissimilarity from natural images are used for target evaluation. Similarity is measured by 3 orthogonal criteria: 1) existence of perspective distortion, 2) the semantic content, and 3) color depth. No data is provided for meta-learning, and target classes are disjoint from the source classes.}
    \label{fig:benchmark}
\end{figure}

% NCFC 11-15: cleaned up writing a bit
Training deep neural networks for visual recognition typically requires a large amount of labelled examples \cite{krizhevsky2012imagenet}. The generalization ability of deep neural networks relies heavily on the size and variations of the dataset used for training. However, collecting sufficient amounts of data for certain classes may be impossible in practice: for example, in dermatology, there are a multitude of instances of rare diseases, or diseases that become rare for particular types of skin \cite{rotemberg2019role,adamson2018machine,fairnessinskin}. Or in other domains such as satellite imagery, there are instances of rare categories such as airplane wreckage. Although individually each situation may not carry heavy cost, as a group across many such conditions and modalities, correct identification is critically important, and remains a significant challenge where access to expertise may be impeded.

Although humans generalize to recognize new categories from few examples in certain circumstances, such as when categories exhibit predictable variations across examples and have reasonable contrast from background \cite{lake2015human,lake2011one}, even humans have trouble recognizing new categories that vary too greatly between examples or differ from prior experience, such as for diagnosis in dermatology, radiology, or other fields \cite{rotemberg2019role}. Because there are many applications where learning must work from few examples, and both machines and humans have difficulty learning in these circumstances, finding new methods to tackle the problem remains a challenging but desirable goal. 
%At this time, there are as many as 245 medical imaging scans per 1,000 persons in the United States \cite{jamami}, and over 126 billion acres on Earth. 

The problem of learning how to categorize classes with very few training examples has been the topic of the ``few-shot learning'' field, and has been the subject of a large body of recent work \cite{li2006one,ravi2016optimization,vinyals2016matching,finn2017model,snell2017prototypical,chen2018a,sung2018learning}. Few-shot learning is typically composed of the following two stages: meta-learning and meta-testing. In the meta-learning stage, there exists an abundance of base category classes on which a system can be trained to learn well under conditions of few-examples within that particular domain. In the meta-testing stage, a set of novel classes consisting of very few examples per class is used to adapt and evaluate the trained model. However, recent work \cite{chen2018a} points out that meta-learning based few-shot learning algorithms underperform compared to traditional pre-training and fine-tuning when there exists a large shift between base and novel class domains. This is a major issue that occurs commonly in practice: by the nature of the problem, collecting data from the same domain for many few-shot classification tasks is difficult. This scenario is referred to as \textit{cross-domain few-shot learning}, to distinguish it from the conventional few-shot learning setting. Although benchmarks for conventional few-shot learning are well established, the cross-domain few-shot learning evaluation benchmarks are still in early stages. All established works in this space have built cross-domain evaluation benchmarks that are limited to natural images \cite{tsengcrossdomain,chen2018a,triantafillou2019meta}. Under these circumstances, useful knowledge may still be effectively transferring across different domains of natural images, implying that methods designed in this setting may not continue to perform well when applied to domains of other types of images, such as industrial natural images, satellite images, or medical images. Currently, no works study this scenario. 

To fill this gap, we propose the Broader Study of Cross-Domain Few-Shot Learning (BSCD-FSL) benchmark (Fig. \ref{fig:benchmark}), which covers a spectrum of image types with varying levels of similarity to natural images. Similarity is defined by 3 orthogonal criteria: 1) whether images contain perspective distortion, 2) the semantic content of images, and 3) color depth. The datasets include agriculture images (natural images, but specific to agriculture industry), satellite (loses perspective distortion), dermatology (loses perspective distortion, and contains different semantic content), and radiological images (different according to all 3 criteria). The performance of existing state-of-art meta-learning methods, transfer learning methods, and methods tailored for cross-domain few-shot learning is then rigorously tested on the proposed benchmark.

%where results reveal that these meta-learning methods perform significantly worse than the standard ``\textit{pre-training and fine-tuning}'' approach. Subsequently, variants of single model fine-tuning techniques are also evaluated, where results demonstrate that no individual method dominates performance across the benchmark, and the relative performance gain with increasing number of shots is greater with transfer methods than compared to meta-learning. In addition, performance across methods also positively correlates with dataset similarity to ImageNet. Further experiments transferring knowledge from multiple pre-trained models, all from disjoint domains than the evaluation benchmark, demonstrates best performance. \\

In summary, the contributions of this paper are itemized as follows:

\begin{itemize}
    % NCFC 11-15: cleaned this up a bit.    
\item We establish a new Broader Study of Cross-Domain Few-Shot Learning (BSCD-FSL) benchmark, consisting of images from a diversity of image types with varying dissimilarity to natural images, according to 1) perspective distortion, 2) the semantic content, and 3) color depth. 
\item Under these conditions, we extensively evaluate the performance of current meta-learning methods, including methods specifically tailored for cross-domain few-shot learning, as well as variants of fine-tuning.
    
\item The results demonstrate that  state-of-art meta-learning methods are outperformed by older meta-learning approaches, and all meta-learning methods underperform in relation to simple fine-tuning by 12.8\% average accuracy. In some cases, meta-learning underperforms even {\em networks with random weights}. 

\item Results also show that accuracy gains for cross-domain few-shot learning methods are lost in this new challenging benchmark. 

\item Finally, we find that accuracy of all methods correlate with the proposed measure of data similarity to natural images, verifying the diversity of the problem representation, and the value of the benchmark towards future research.

\end{itemize}

We believe this work will help the community understand what methods are most effective in practice, and help drive further advances that can more quickly yield benefit for real-world applications.

\section{Related Work}
\subsubsection{Few-shot learning}
Few-shot learning \cite{lake2015human,vinyals2016matching,lake2011one} is an increasingly important topic in machine learning. Many few-shot methods have been proposed, including meta-learning, generative and augmentation approaches, semi-supervised methods, and transfer learning. 

Meta-learning methods aim to learn models that can be quickly adapted using a few examples \cite{vinyals2016matching,finn2017model,snell2017prototypical,sung2018learning,lee2019meta}. MatchingNet \cite{vinyals2016matching} learns an embedding that can map an unlabelled example to its label using a small number of labelled examples, while MAML \cite{finn2017model} aims at learning good initialization parameters that can be quickly adapted to a new task. In ProtoNet \cite{snell2017prototypical}, the goal is to learn a metric space in which classification can be conducted by calculating distances to prototype representations of each class. RelationNet \cite{sung2018learning} targets learning a deep distance metric to compare a small number of images. More recently, MetaOpt \cite{lee2019meta} learns feature embeddings that can generalize well under a linear classification rule for novel categories.

The generative and augmentation based family of approaches learn to generate more samples from few examples available for training in a given few-shot learning task. These methods include applying augmentation strategies learned from data \cite{lim2019a}, synthesizing new data from few examples using a generative model, or using external data for obtaining additional examples that facilitate learning on a given few shot task. In \cite{Hariharan2017,Schwartz2018} the intra-class relations between pairs of instances of reference categories are modeled in feature space, and then this information is transferred to the novel category instances to generate additional examples in that same feature space. In \cite{Wang2018}, a generator sub-net is added to a classifier network and is trained to synthesize new examples on the fly in order to improve the classifier performance when being fine-tuned on a novel (few-shot) task. In \cite{Reed2018}, a few-shot class density estimation is performed with an auto-regressive model, combined with an attention mechanism, where examples are synthesized by a sequential process. In \cite{Chen2018,Schwartz2019,Yu2017} label and attribute semantics are used as additional information for training an example synthesis network.

In some situations there exists additional unlabeled data accompanying the few-shot task. In the semi-supervised few-shot learning \cite{Li2019a,Ren2018,projective2019,Liu2019x,saenko} the unlabeled data comes in addition to the support set and is assumed to have a similar distribution to the target classes (although some unrelated samples noise is also allowed). In LST \cite{Li2019a}, self-labeling and soft attention are used on the unlabeled samples intermittently with fine-tuning on the labeled and self-labeled data. Similarly to LST, \cite{Ren2018} updates the class prototypes using k-means like iterations initialized from the PN prototypes. In \cite{projective2019}, unlabeled examples are used through soft-label propagation. In \cite{Garcia2017,Liu2019x,Kim2019}, graph neural networks are used for sharing information between labeled and unlabeled examples in semi-supervised \cite{Garcia2017,Liu2019x} and transductive \cite{Kim2019} FSL setting. Notably, in \cite{Liu2019x} a Graph Construction network is used to predict the task specific graph for propagating labels between samples of semi-supervised FSL task.

Transfer learning \cite{pan2009survey} is based on the idea of reusing features learned from the base classes for the novel classes, and is conducted mainly by fine-tuning, which adjusts a pre-trained model from a source task to a target task. Yosinski et al. \cite{yosinski2014transferable} conducted extensive experiments to investigate the transfer utility of pre-trained deep neural networks. In \cite{kornblith2018better}, the authors investigated whether higher performing ImageNet models transfer better to new tasks. Ge et al. \cite{ge2017borrowing} proposed a selective joint fine-tuning method for improving the performance of models with a limited amount training data. In \cite{guo2019spottune}, the authors proposed an adaptive fine-tuning scheme to decide which layers of the pre-trained network should be fine-tuned. Finally, in \cite{dhillonfsbaseline}, the authors found that simple transductive fine-tuning beats all prior state-of-art meta-learning approaches.

\iffalse 
\begin{figure*}
\centering
\begin{subfigure}{.5\textwidth}
  \centering
  \includegraphics[width=0.4\linewidth]{./figs/imagenet.png}
  \caption{Source domain: \textit{ImageNet}}
  \label{fig:imagenet}
\end{subfigure}%
\begin{subfigure}{.5\textwidth}
  \includegraphics[width=0.8\linewidth]{./figs/cross.jpg}
  \caption{The proposed cross-domain few-shot learning benchmark includes \textit{ChestX}, \textit{ISIC2018}, \textit{EuroSAT} and \textit{CropDiseases}.}
  \label{fig:benchmark}
\end{subfigure}
\caption{Cross-domain few-shot learning: the model is trained on a source domain consisting of natural images such as ImageNet and used for a new task consisting of a few examples per class sampled from a different target domain. }
\label{fig:test}
\end{figure*}
\fi

%\textbf{Few-shot Learning Benchmarks.}
Common to all few-shot learning methods is the assumption that {\em base classes and novel classes are from the same domain}. The current benchmarks for evaluation are miniImageNet \cite{vinyals2016matching}, CUB \cite{wah2011caltech}, Omniglot \cite{lake2011one}, CIFAR-FS \cite{bertinetto2018meta} and tieredImageNet \cite{ren2018meta}.  In \cite{triantafillou2019meta}, the authors proposed Meta-Dataset, which is a newer benchmark for training and evaluating few-shot learning algorithms that includes a greater diversity of image content. Although this benchmark is more broad than prior works, the included datasets are still limited to {\em natural images}, and both the base classes and novel classes are from the {\em same domain}. Recently, \cite{requeima2019fast} proposes a successful meta-learning approach based on conditional neural process on the MetaDataset benchmark.

%Arguably, the current few-shot learning benchmarks do not reflect the reality of few-shot learning applications where meta-learning data in domain is commonly not available. 

\subsubsection{Domain Adaptation}
There is a long history of research in domain adaptation techniques, which aim at transferring knowledge from one or multiple source domains to a target domain with a different data distribution. 
Early methods have generally relied on the adaptation of {\em shallow} classification models, using techniques such as instance re-weighting \cite{dudik2006correcting} and model parameter adaptation \cite{yang2007cross}. More recently, many methods have been proposed to address the problem of domain adaptation using deep neural networks, including discrepancy-based methods, designed to align marginal distributions between the domains \cite{long2017deep,sun2016return,kang2019contrastive,kumar2018co}, adversarial-based approaches, which rely on a domain discriminator to encourage domain-independent feature learning \cite{tzeng2017adversarial,ganin2016domain}, and reconstruction-based techniques, which generally use encoder-decoder models or GANs to reconstruct data in the new domain \cite{bousmalis2016domain,zhu2017unpaired,hoffman2017cycada}. All these approaches, however, consider the case that the training and test sets have the {\em same classes}. One work considers the scenario where some classes may be disjoint, but still requires class overlap for successful alignment \cite{saenko2}. In contrast, we study the problem of cross-domain few-shot learning, where the source and target domains have completely {\em disjoint label sets}. 
%For domain adaptation theory \cite{ben2010theory}, the idea is to adapt a model from a source dataset to a target dataset, for a fixed set of labels. {\em TO DO: Cover relevant literature.}

%Different from the traditional domain adaptation setting, the label space of source domain and target domain in cross-domain few-shot learning is disjoint. One work studies the intersection of domain adaptation and one-shot learning, but requires the availability of unlabeled data \cite{dongxing}, which may not be available in practice.  

\subsubsection{Cross-domain Few-shot Learning}

In cross-domain few-shot learning, base and novel classes are both drawn from different domains, and the class label sets are disjoint. Recent works on cross-domain few-shot learning include analysis of existing meta-learning approaches in the cross-domain setting \cite{chen2018a}, specialized methods using feature-wise transform to encourage learning representations with improved ability to generalize \cite{tsengcrossdomain}, and works studying cross-domain few-shot learning constrained to the setting of images of items in museum galleries \cite{openmic}. Common to all  these prior works is that they limit the cross-domain setting to the realm of {\em natural images}, which still retain a high degree of visual similarity, and do not capture the broader spectrum of image types encountered in practice, such as industrial, aerial, and medical images, where cross-domain few-shot learning techniques are in high demand. 

\section{Proposed Benchmark}
In this section, we introduce the  Broader Study of Cross-Domain Few-Shot Learning (BSCD-FSL) benchmark, which includes data from CropDiseases \cite{mohanty2016using}, EuroSAT \cite{helber2019eurosat}, ISIC2018 \cite{tschandl2018ham10000,codella2019skin}, and ChestX \cite{wang2017chestx} datasets. These datasets cover plant disease images, satellite images, dermoscopic images of skin lesions, and X-ray images, respectively.  The selected datasets reflect well-curated real-world use cases for few-shot learning. In addition, collecting enough examples from above domains is often difficult, expensive, or in some cases not possible. Image similarity to natural images is measured by 3 orthogonal criteria: 1) existeence of perspective distortion, 2) the semantic data content, and 3) color depth. According to this criteria, the datasets demonstrate the following spectrum of image types: 1) CropDiseases images are natural images, but are very specialized (similar to existing cross-domain few-shot setting, but specific to agriculture industry), 2) EuroSAT images are less similar as they have lost perspective distortion, but are still color images of natural scenes, 3) ISIC2018 images are even less similar as they have lost perspective distortion and no longer represent natural scenes, and 4) ChestX images are the most dissimilar as they have lost perspective distortion, do not represent natural scenes, and have lost 2 color channels. Example images from ImageNet and the proposed benchmark datasets are shown in Figure \ref{fig:benchmark}. 

Having a few-shot learning model trained on a source domain such as ImageNet \cite{deng2009imagenet} that can generalize to domains such as these, is highly desirable, as it enables effective learning for rare categories in new types of images, which has previously not been studied in detail. 

\section{Cross-Domain Few-Shot Learning Formulation}

The cross domain few-shot learning problem can be formalized as follows. We define a \textit{domain} as a joint distribution $P$ over input space $\mathcal{X}$ and label space $\mathcal{Y}$. The marginal distribution of $\mathcal{X}$ is denoted as $P_\mathcal{X}$. We use the pair $(x, y)$ to denote a sample $x$ and the corresponding label $y$ from the joint distribution $P$. For a model $f_\theta$ : $\mathcal{X}$ $\rightarrow$ $\mathcal{Y}$ with parameter $\theta$ and a loss function $\ell$, the expected error is defined as,
\begin{equation}
    \epsilon(f_\theta) = E_{(x, y) \sim P} [\ell(f_\theta(x), y)]
\end{equation}

In cross-domain few-shot learning, we have a source domain $(\mathcal{X}_s, \mathcal{Y}_s)$ and a target domain $(\mathcal{X}_t, \mathcal{Y}_t)$ with joint distribution $P_s$ and $P_t$ respectively,  $P_{\mathcal{X}_s} \neq P_{\mathcal{X}_t}$, and $\mathcal{Y}_s$  is disjoint from $\mathcal{Y}_t$. The base classes data are sampled from the source domain and the novel classes data are sampled from the target domain. During the training or meta-training stage, the model $f_\theta$ is trained (or meta-trained) on the base classes data. During testing (or meta-testing) stage, the model is presented with a \textit{support set} $S = \{x_i, y_i\}_{i=1}^{K \times N}$ consisting of $N$ examples from $K$ novel classes. This configuration is referred to as ``$K$-way $N$-shot'' few-shot learning, as the support set has $K$ novel classes and each novel class has $N$ training examples. After the model is adapted to the support set, a \textit{query set} from novel classes is used to evaluate the model performance.

\section{Evaluated Methods for Cross-Domain Few-Shot Learning}
\label{sec:methods}
In this section, we describe the few-shot learning algorithms that will be evaluated on our proposed benchmark.
%We study meta-learning based methods and transfer learning based methods. For meta-learning methods, we additionally study algorithm modifications specifically designed for cross-domain few-shot learning \cite{tsengcrossdomain}. For transfer learning methods, we also study the effect of different type of classifiers, and whether transferring from multiple models pre-trained on different natural image datasets can generally boost performance. 

%\subsection {Single Model Techniques}   
\subsection{Meta-learning based methods}

\subsubsection{Single Domain Methods}
Meta-learning \cite{finn2017model,ravi2016optimization}, or learning to learn, aims at learning task-agnostic knowledge in order to efficiently learn on new tasks. Each task $\mathcal{T}_i$ is assumed to be drawn from a fixed distribution, $\mathcal{T}_i \sim P(\mathcal{T})$. Specially, in few-shot learning, each task $\mathcal{T}_i$ is a small dataset $D_i \coloneqq \{x_j, y_j\}_{j=1}^{K \times N}$. $P_s(\mathcal{T})$ and $P_t(\mathcal{T})$ are used to denote the task distribution of the source (base) classes data and target (novel) classes data respectively. During the meta-training stage, the model is trained on $T$ tasks $\{\mathcal{T}_i\}_{i=1}^T$ which are sampled independently from $P_s(\mathcal{T})$. During the meta-testing stage, the model is expected to be quickly adapted to a new task ${T}_j \sim P_t(\mathcal{T})$.

Meta-learning methods differ in their way of learning the parameter of the initial model $f_\theta$ on the base classes data. In MatchingNet \cite{vinyals2016matching}, the goal is to learn a model $f_\theta$ that can map an unlabelled example $\hat{x}$ to its label $\hat{y}$ using a small labelled set $D_i \coloneqq \{x_j, y_j\}_{j=1}^{K \times N}$ as $\hat{y} = \sum_{j=1}^{K \times N}a_\theta(\hat{x}, x_j)y_j$, where $a_\theta$ is an attention kernel which leverages $f_\theta$ to compute the distance between the unlabelled example $\hat{x}$ and the labelled example $x_j$, and $y_j$ is the one-hot representation of the label. In contrast, MAML \cite{finn2017model} aims at learning an initial parameter $\theta$ that can be quickly adapted to a new task. This is achieved by updating the model parameter via a two-stage optimization process. ProtoNet \cite{snell2017prototypical} represents each class $k$ with the mean vector of embedded support examples as $c_k = \frac{1}{N} \sum_{j=1}^{N}f_\theta(x_j)$. Classification is then conducted by calculating distance of the example to the prototype representations of each class. In RelationNet \cite{sung2018learning} the metric of the nearest neighbor classifier is meta-learned using a Siamese Networks trained for optimal comparison between query and support samples. More recently, MetaOpt \cite{lee2019meta} employs convex base learners and aims at learning feature embeddings that generalize well under a linear classification rule for novel categories. All the existing meta-learning methods implicitly assume that $P_s(\mathcal{T})$ = $P_t(\mathcal{T})$ so the task-agnostic knowledge learned in the meta-training stage can be leveraged for fast learning on novel classes. However, in cross-domain few-shot learning $P_s(\mathcal{T})$ $\neq$ $P_t(\mathcal{T})$ which poses severe challenges for current meta-learning methods.

\subsubsection{Cross-Domain Methods}

Only few methods specifically tailored to learning in the condition of cross-domain few-shot learning have been previously explored, including feaure-wise transform (FWT) \cite{tsengcrossdomain}, and Adversarial Domain Adaptation with Reinforced Sample (ADA-RSS)
Selection \cite{dongxing}. Since the problem setting of ADA-RSS requires the existence of unlabelled data in the target domain, we study FWT alone. 

FWT is a model agnostic approach that adds a feature-wise transform layer to pre-trained models to learn scale and shift parameters from a collection of several dataset domains, or use parameters empirically determined from a single dataset domain. Both approaches have been previously found to improve performance. Since our benchmark is focused on ImageNet as the single source domain, we focus on the single data domain approach. The method is studied in combination with all meta-learning algorithms described in the prior section.

\subsection{Transfer learning based methods}
An alternative way to tackle the problem of few-shot learning is based on transfer learning, where an initial model $f_\theta$ is trained on the base classes data in a standard supervised learning way and reused on the novel classes. There are several options to realize the idea of transfer learning for few-shot learning:

%Those studied in this work are outlined as follows:

%Previous works on transfer learning for few-shot learning \cite{snell2017prototypical,chen2018a} simply freeze the pre-trained model and use it as a fixed feature extractor. 
%While it was pointed out that fine-tuning the pre-trained model on the novel classes data would lead to overfitting in the limited-data regime \cite{snell2017prototypical,sung2018learning}, our results show that the conclusion does not hold in cross-domain few-shot learning.

\subsubsection{Single Model Methods}
 In this paper, we extensively evaluate the following commonly variants of single model fine-tuning:

\begin{itemize}
% this is the commonly used transfer learning strategy for few-shot learning which 
    %\setlength\itemsep{0.7em}
    \item \textit{Fixed feature extractor (Fixed)}: simply leverage the pre-trained model as a fixed feature extractor. 
    
    \item \textit{Fine-tuning all layers} (\textit{Ft All}): adjusts all the pre-trained parameters on the new task with standard
    supervised learning. 
    
    \item \textit{Fine-tuning last\--k (Ft Last-k)}: only the last $k$ layers of the pre-trained model are optimized for the new task. In the paper, we consider Fine-tuning last\--1,  Fine-tuning last\--2, Fine-tuning last\--3.
    
    \item \textit{Transductive fine-tuning (Transductive Ft)}: in transductive fine-tuning, the statistics of the query images are used via batch normalization \cite{dhillonfsbaseline}.  \cite{nichol2018first}.
\end{itemize}

In addition, we compare these single model transfer learning techniques against a baseline of an embedding formed by a randomly initialized network (termed {\em Random}) to contrast against a fixed feature vector that has no pre-training. All the variants of single model fine-tuning are based on linear classifier but differ in their approach to fine-tune the single model feature extractor.

Another line of work for few-shot learning uses a broader variety of classifiers for transfer learning. For example, recent works show that mean-centroid classifier and cosine-similarity based classifier are more effective than linear classifier for few-shot learning \cite{mensink2013distance,chen2018a}. Therefore we study these two variations as well.

\paragraph{Mean-centroid classifier.}
The mean-centroid classifier is inspired from ProtoNet \cite{snell2017prototypical}. Given the pre-trained model $f_{\theta}$ and a support set $S = \{x_i, y_i\}_{i=1}^{K \times N}$, where $K$ is the number of novel classes and $N$ is the number of images per class. The class prototypes are computed in the same way as in ProtoNet. Then the likelihood of an unlabelled example $\hat{x}$ belongs to class $k$ is computed as,
\begin{equation}
    p(y=k | \hat{x}) = \frac{\textnormal{exp}(-d(f_{\theta}, c_k ))}{\sum_{l=1}^K \textnormal{exp}(-d(f_{\theta}, c_l))}
\end{equation}
where $d()$ is a distance function. In the experiments, we use negative cosine similarity. %Other distance functions such as Euclidean distance can also be used. 
Different from ProtoNet, $f_\theta$ is pretrained on the base classes data in a standard supervised learning way.

\paragraph{Cosine-similarity based classifier.} 
In cosine-similarity based classifier, instead of directly computing the class prototypes using the pre-trained model, each class $k$ is represented as a $d$-dimension weight vector $\mathbf{w}_k$ which is initialized randomly. For each unlabeled example $\hat{x}_i$, the cosine similarity to each weight vector is computed as $c_{i,k} = \frac{f_{\theta}(\hat{x}_i)^T \mathbf{w}_k}{\lVert f_{\theta}(\hat{x}_i) \rVert \lVert \mathbf{w}_k \rVert }$. The predictive probability of the example $\hat{x}_i$ belongs to class $k$ is computed by normalizing the cosine similarity with a softmax function. Intuitively, the weight vector $\mathbf{w}_k$ can be thought as the prototype of class $k$.

\subsubsection{Transfer from Multiple Pre-trained Models}
In this section, we describe a straightforward method that utilizes multiple models pre-trained on source domains of natural images similar to ImageNet. Note that all domains are still disjoint from the target datasets for the cross-domain few-shot learning setting. The purpose is to measure how much performance may improve by utilizing an ensemble of models trained from data that is different from the target domain. The described method requires no change to how models are trained and is an off-the-shelf solution to leverage existing pre-trained models for cross-domain few-shot learning, without requiring access to the source datasets.

%Unlike previous works on ensemble methods for few-shot learning \cite{dvornik2019diversity} that train diverse models on the same source dataset, the described method requires no change to how models are trained and is an off-the-shelf solution to leverage existing pre-trained models for cross-domain few-shot learning, without requiring access to the source datasets.

Assume we have a library of $C$ pre-trained models $\{M_c\}_{c=1}^{C}$ which are trained on various datasets in a standard way. We denote the layers of all pre-trained models as a set $F$. Given a support set $S = \{x_i, y_i\}_{i=1}^{K \times N}$ where $(x_i, y_i) \sim P_t$, our goal is to find a subset $I$ of the layers to generate a feature vector for each example in order to achieve the lowest test error. Mathematically, 
 
\begin{equation}
\label{selection}
    \argmin_{I \subseteq F} \,_{(x, y) \sim\ P_t} \ell ( f_s(  T (\{l(x): l \in I \}), y)
\end{equation}

\noindent where $\ell$ is a loss function, $T()$ is a function which concatenates a set of feature vectors, $l$ is one particular layer in the set $I$, and $f_s$ is a linear classifier. Practically, for feature vectors $l$ coming from inner layers which are three-dimensional, we convert them to one-dimensional vectors by using Global Average Pooling. Since Eq. \ref{selection} is intractable generally, we instead adopt a two-stage greedy selection method, called \textit{Incremental Multi-model Selection}, to iteratively find the best subset of layers for a given support $S$. 

In the first stage, for each pre-trained model, we a train linear classifier on the feature vector generated by each layer individually and select the corresponding layer which achieves the lowest average error using five-fold cross-validation on the support set $S$. Essentially, the goal of the first stage is to find the most effective layer of each pre-trained model given the task in order to reduce the search space and mitigate risk of overfitting. For convenience, we denote the layers selected in the first selection stage as set $I_1$. In the second stage, we greedily add the layers in $I_1$ into the set $I$ following a similar cross-validation procedure. First, we add the layer in $I_1$ into $I$ which achieves the lowest cross-validation error. Then we iterate over $I_1$, and add each remaining layer into $I$ if the cross-validation error is reduced when the new layer is added. Finally, we concatenate the feature vector generated by each layer in set $I$ and train the final linear classifier. Please see Algorithm \ref{alg: ims} in Appendix for further details.

\begin{table*}[!t!]
\centering
\scalebox{0.7}{
\begin{tabular}{c c c c | c c c}
    \hline
    \hline
\textbf{Methods} &\multicolumn{3}{c}{\textbf{ChestX}}& \multicolumn{3}{c}{\textbf{ISIC}}  \\
    \cline{2-7}
    & 5-way 5-shot &5-way 20-shot&5-way 50-shot&5-way 5-shot &5-way 20-shot &5-way 50-shot \\
    \cline{2-7}
\emph{MatchingNet}   & 22.40\% $\pm$ 0.7\% & 23.61\% $\pm$ 0.86\% &22.12\% $\pm$ 0.88\% & 36.74\% $\pm$ 0.53\%& 45.72\% $\pm$ 0.53\%&  54.58\% $\pm$ 0.65\%\\

\emph{MatchingNet+FWT}   & 21.26\% $\pm$ 0.31\%& 23.23\% $\pm$ 0.37\%  & 23.01\% $\pm$ 0.34\% &  30.40\% $\pm$ 0.48\% & 32.01\% $\pm$ 0.48\% & 33.17\% $\pm$ 0.43\% \\

\emph{MAML}  & 23.48\% $\pm$ 0.96\% & 27.53\% $\pm$ 0.43\%  & - &\textbf{40.13\% $\pm$ 0.58\%} &\textbf{52.36\% $\pm$ 0.57\%} & - \\
  
\emph{ProtoNet}   & \textbf{24.05\% $\pm$ 1.01\%}& \textbf{28.21\% $\pm$ 1.15\%} &29.32\% $\pm$ 1.12\% & 39.57\% $\pm$ 0.57\%&49.50\% $\pm$ 0.55\%&51.99\% $\pm$ 0.52\%\\

\emph{ProtoNet+FWT}   & 23.77\% $\pm$ 0.42\% & 26.87\% $\pm$ 0.43\% & \textbf{30.12\% $\pm$ 0.46\%} &  38.87\% $\pm$ 0.52\% & 43.78\% $\pm$ 0.47\% & 49.84\% $\pm$ 0.51\%\\

\emph{RelationNet}   & 22.96\% $\pm$ 0.88\%&26.63\% $\pm$ 0.92\%&28.45\% $\pm$ 1.20\% & 39.41\% $\pm$ 0.58\% & 41.77\% $\pm$ 0.49\% &49.32\% $\pm$ 0.51\% \\

\emph{RelationNet+FWT}   &22.74\% $\pm$ 0.40\% & 26.75\% $\pm$ 0.41\%& 27.56\% $\pm$ 0.40\% &35.54\% $\pm$ 0.55\% & 43.31\% $\pm$ 0.51\%& 46.38\% $\pm$ 0.53\%\\

\emph{MetaOpt}  &22.53\% $\pm$ 0.91\% &25.53\% $\pm$ 1.02\%& 29.35\% $\pm$ 0.99\%&36.28\% $\pm$ 0.50\% &49.42\% $\pm$ 0.60\%&\textbf{54.80\% $\pm$ 0.54\% }\\
    \hline
\end{tabular}}
\end{table*}
\begin{table*}[!t!]
\centering
\scalebox{0.7}{
\begin{tabular}{c c c c | c c c}
    \hline
    \hline
\textbf{Methods} &\multicolumn{3}{c}{\textbf{EuroSAT}}& \multicolumn{3}{c}{\textbf{CropDiseases}}  \\
    \cline{2-7}
    & 5-way 5-shot &5-way 20-shot&5-way 50-shot&5-way 5-shot &5-way 20-shot &5-way 50-shot \\
    \cline{2-7}
\emph{MatchingNet}   & 64.45\% $\pm$ 0.63\%&77.10\% $\pm$ 0.57\%&54.44\% $\pm$ 0.67\% &66.39\% $\pm$ 0.78\%  & 76.38\% $\pm$ 0.67\%& 58.53\% $\pm$ 0.73\%\\

\emph{MatchingNet+FWT}   & 56.04\% $\pm$ 0.65\%&  63.38\% $\pm$ 0.69\% & 62.75\% $\pm$ 0.76\% & 62.74\% $\pm$ 0.90\% & 74.90\% $\pm$ 0.71\% & 75.68\% $\pm$ 0.78\%\\

\emph{MAML}  & 71.70\% $\pm$ 0.72\%&81.95\% $\pm$ 0.55\%& -& 78.05\% $\pm$ 0.68\%&\textbf{89.75\% $\pm$ 0.42\%} &-\\
  
\emph{ProtoNet}   &\textbf{73.29\% $\pm$ 0.71\%} &  \textbf{82.27\% $\pm$ 0.57\% }&80.48\% $\pm$ 0.57\% & \textbf{79.72\% $\pm$ 0.67\%} &88.15\% $\pm$ 0.51\% &90.81\% $\pm$ 0.43\% \\

\emph{ProtoNet+FWT}   & 67.34\% $\pm$ 0.76\%& 75.74\% $\pm$ 0.70\%  & 78.64\% $\pm$ 0.57\%& 72.72\% $\pm$ 0.70\% &85.82\% $\pm$ 0.51\%&87.17\% $\pm$ 0.50\%\\

\emph{RelationNet}   & 61.31\% $\pm$ 0.72\%  & 74.43\% $\pm$ 0.66\%& 74.91\% $\pm$ 0.58\% &  68.99\% $\pm$ 0.75\% &80.45\% $\pm$ 0.64\%&85.08\% $\pm$ 0.53\%\\

\emph{RelationNet+FWT}   & 61.16\% $\pm$ 0.70\% & 69.40\% $\pm$ 0.64\% & 73.84\% $\pm$ 0.60\% & 64.91\% $\pm$ 0.79\%  & 78.43\% $\pm$ 0.59\%& 81.14\% $\pm$ 0.56\%\\

\emph{MetaOpt}  & 64.44\% $\pm$ 0.73\% &79.19\% $\pm$ 0.62\%&\textbf{83.62\% $\pm$ 0.58\% } & 68.41\% $\pm$ 0.73\%&82.89\% $\pm$ 0.54\%&\textbf{91.76\% $\pm$ 0.38\%}\\
    \hline
\end{tabular}}
\caption{The results of meta-learning methods on the proposed benchmark.}
\label{tab:meta}
\end{table*}

\section{Evaluation Setup}
\label{sec:setup}
For meta-learning methods, we meta-train all meta-learning methods on the base classes of miniImageNet \cite{vinyals2016matching} and meta-test the trained models on each dataset of the proposed benchmark. For transfer learning methods, we train the pre-trained model on base classes of miniImageNet. For transferring from multiple pre-trained models, we use a maximum of five pre-trained models, trained on miniImagenet, CIFAR100 \cite{krizhevsky2009learning}, DTD \cite{cimpoi2014describing}, CUB \cite{WelinderEtal2010}, Caltech256 \cite{griffin2007caltech}, respectively. On all experiments we consider 5-way 5-shot, 5-way 20-shot, 5-way 50-shot. For all cases, the test (query) set has 15 images per class. All experiments are performed with ResNet-10 \cite{he2016deep} for fair comparison. For each evaluation, we use the same 600 randomly sampled few-shot episodes (for consistency), and report the average accuracy and $95\%$ confidence interval.  

 During the training (meta-training) stage, models used for transfer learning and meta-learning models are both trained for 400 epochs with Adam optimizer. The learning rate is set to 0.001. During testing (meta-testing), both transfer learning methods and those meta-learning methods that require adaptation on the support set of the test episodes (MAML, RelationNet, etc.) use SGD with momentum. The learning rate is 0.01 and the momentum rate is 0.9. All variants of fine-tuning methods are trained for 100 epochs. For feature-wise transformation \cite{tsengcrossdomain}, we adopt the recommended hyperparameters in the original paper for meta-training from one source domain . In the training or meta-training stage, we apply standard data augmentation including random crop, random flip, and color jitter.

In the cross-domain few-shot learning setting, since the source domain and target domain are drastically different, it may not be appropriate to use the source domain data for hyperparameter tuning or validation. Therefore, we leave the question of how to determine the best hyperparameters in the  cross-domain few-shot learning as future work. One simple strategy is to use the test set or validation set of the source domain data for hyperparameter tuning. More sophisticated methods may use datasets that are similar to the target domain data.

\section{Experimental Results}

%Results are discussed according to method categories as described in Section \ref{sec:methods}. First, results from some recent state-of-art and classic meta-learning methods are presented in Section \ref{subsec:meta}. Next, transfer learning is evaluated and analyzed in Section \ref{sec:transfer}, including single model transfer followed by multi-model transfer. Finally, a succinct best-in-category comparison is presented in Section \ref{subsec:inter}. 

\subsection{Meta-learning based results}
\label{subsec:meta}
Table \ref{tab:meta} show the results on the proposed benchmark of meta-learning, for each dataset, method, and shot level in the benchmark. Across all datasets and shot levels, the average accuracies (and 95\% confidence internals) are 50.21\% (0.70) for MatchingNet, 46.55\% (0.58) for MatchingNet+FWT, 38.75\% (0.41) for MAML, 59.78\% (0.70) for ProtoNet, 56.72\% (0.55) for ProtoNet+FWT, 54.48\% (0.71) for RelationNet, 52.6\% (0.56) for RelationNet+FWT, and 57.35\% (0.68) for MetaOpt. The performance of MAML was impacted by its inability to scale to larger shot levels due to memory overflow. Methods paired with Feature-Wise Transform are marked with ``+FWT''.  

What is immediately apparent from Table \ref{tab:meta}, is that the prior state-of-art MetaOptNet is no longer state-of-art, as it is outperformed by ProtoNet. In addition, methods designed specifically for cross-domain few-shot learning lead to consistent performance degradation in this new challenging benchmark. Finally, performance in general strongly positively correlates to the dataset's similarity to ImageNet, confirming that the benchmark's intentional design allows us to investigate few-shot learning in a spectrum of cross-domain difficulties.

\begin{table*}[!t]
\centering
\scalebox{0.7}{
\begin{tabular}{c c c c | c c c}
    \hline
    \hline
\textbf{Methods} &\multicolumn{3}{c}{\textbf{ChestX}}& \multicolumn{3}{c}{\textbf{ISIC}}  \\
    \cline{2-7}
    & 5-way 5-shot &5-way 20-shot&5-way 50-shot&5-way 5-shot &5-way 20-shot &5-way 50-shot \\
    \cline{2-7}

\emph{Random} &21.80\% $\pm$ 1.03\%& 25.69\% $\pm$ 0.95\%& 26.19\% $\pm$ 0.94\%& 37.91\% $\pm$ 1.39\% & 47.24\% $\pm$ 1.50\%&50.85\% $\pm$ 1.37\%\\

\emph{Fixed}  & 25.35\% $\pm$ 0.96\% &30.83\% $\pm$ 1.05\%& 36.04\% $\pm$ 0.46\%& 43.56\% $\pm$ 0.60\% &52.78\% $\pm$ 0.58\%&57.34\% $\pm$ 0.56\%\\
  
\emph{Ft All} & 25.97\% $\pm$ 0.41\%&  31.32\% $\pm$ 0.45\%&  35.49\% $\pm$ 0.45\%& 48.11\% $\pm$ 0.64\% &59.31\% $\pm$ 0.48\%&66.48\% $\pm$ 0.56\%\\

\emph{Ft Last-1}   & 25.96\% $\pm$ 0.46\% &  \textbf{31.63\% $\pm$ 0.49\%} & 37.03\% $\pm$ 0.50\% & 47.20\% $\pm$ 0.45\% & 59.95\% $\pm$ 0.45\% &  65.04\% $\pm$ 0.47\%\\

\emph{Ft Last-2}   &  \textbf{26.79\% $\pm$ 0.59\%} & 30.95\% $\pm$ 0.61\% &  36.24\% $\pm$ 0.62\% & 47.64\% $\pm$ 0.44\% & 59.87\% $\pm$ 0.35\%& 66.07\% $\pm$ 0.45\%\\

\emph{Ft Last-3}   &  25.17\% $\pm$ 0.56\% & 30.92\% $\pm$ 0.89\%& \textbf{37.27\% $\pm$ 0.64\% }& 48.05\% $\pm$ 0.55\% & 60.20\% $\pm$ 0.33\% &66.21\% $\pm$ 0.52\% \\

\emph{Transductive Ft} & 26.09\% $\pm$ 0.96\%&31.01\% $\pm$ 0.59\%& 36.79\% $\pm$ 0.53\%& \textbf{49.68\% $\pm$ 0.36\%}&\textbf{61.09\% $\pm$ 0.44\%}&\textbf{67.20\% $\pm$ 0.59\%} \\
    \hline
\end{tabular}}
\end{table*}

\begin{table*}[!t]
\centering
\scalebox{0.7}{
\begin{tabular}{c c c c | c c c}
    \hline
    \hline
\textbf{Methods} &\multicolumn{3}{c}{\textbf{EuroSAT}}& \multicolumn{3}{c}{\textbf{CropDiseases}}  \\
    \cline{2-7}
    & 5-way 5-shot &5-way 20-shot&5-way 50-shot&5-way 5-shot &5-way 20-shot &5-way 50-shot \\
    \cline{2-7}
\emph{Random} &  58.00\% $\pm$ 2.01\%&68.93\% $\pm$ 1.47\%& 71.65\% $\pm$ 1.47\%& 69.68\% $\pm$ 1.72\%& 83.41\% $\pm$ 1.25\% &86.56\% $\pm$ 1.42\%\\

\emph{Fixed}  & 75.69\% $\pm$ 0.66\%&84.13\% $\pm$ 0.52\% &86.62\% $\pm$ 0.47\% & 87.48\% $\pm$ 0.58\%&94.45\% $\pm$ 0.36\%&96.62\% $\pm$ 0.25\%\\
  
\emph{Ft All}  &79.08\% $\pm$ 0.61\% &87.64\% $\pm$ 0.47\%& 90.89\% $\pm$ 0.36\% & 89.25\% $\pm$ 0.51\%& 95.51\% $\pm$ 0.31\%&97.68\% $\pm$ 0.21\%\\

\emph{Ft Last-1}   &80.45\% $\pm$ 0.54\% & 87.92\% $\pm$ 0.44\%& 91.41\% $\pm$ 0.46\% & 88.72\% $\pm$ 0.53\%& 95.76\% $\pm$ 0.65\% & \textbf{97.87\% $\pm$ 0.48\%}\\

\emph{Ft Last-2}   & 79.57\% $\pm$ 0.51\% & 87.67\% $\pm$ 0.46\%& 90.93\% $\pm$ 0.45\%  &  88.07\% $\pm$ 0.56\% & 95.68\% $\pm$ 0.76\% & 97.64\% $\pm$ 0.59\%\\

\emph{Ft Last-3}   & 78.04\% $\pm$ 0.77\% &  87.52\% $\pm$ 0.53\% & 90.83\% $\pm$ 0.42\% & 89.11\% $\pm$ 0.47\%&  95.31\% $\pm$ 0.7\% &  97.45\% $\pm$ 0.46\% \\

\emph{Transductive Ft} &  \textbf{81.76\% $\pm$ 0.48\%}&\textbf{87.97\% $\pm$ 0.42\%}&\textbf{92.00\% $\pm$ 0.56\%} & \textbf{90.64\% $\pm$ 0.54\% }&\textbf{95.91\% $\pm$ 0.72\%}& 97.48\% $\pm$ 0.56\% \\
    \hline
\end{tabular}}
\caption{The results of different variants of single model fine-tuning on the proposed benchmark.}
\label{tab:finetune}
\end{table*}

\subsection{Transfer learning based results}
\label{sec:transfer}
\subsubsection{Single model results}
\label{subsec:single}
Table \ref{tab:finetune} show the results on the proposed benchmark of various single model transfer learning methods. Across all datasets and shot levels, the average accuracies (and 95\% confidence internals) are 53.99\% (1.38) for random embedding, 64.24 (0.59) for fixed feature embedding, 67.23\% (0.46) for fine-tuning all layers, 67.41\% (0.49) for fine-tuning the last 1 layer, 67.26\% (0.53) for fine-tuning the last 2 layers, 67.17\% (0.58) for fine-tuning the last 3 layers, and 68.14\% (0.56) for transductive fine-tuning.  From these results, several observations can be made. The first observation is that, although meta-learning methods have been previously shown to achieve higher performance than transfer learning in the standard few-shot learning setting \cite{vinyals2016matching,chen2018a}, in the cross-domain few-shot learning setting this situation is reversed: meta-learning methods significantly underperform simple fine-tuning methods. In fact, \emph{ MatchingNet performs worse than a randomly generated fixed embedding}. A possible explanation is that meta-learning methods are fitting the task distribution on the base class data, improving performance in that circumstance, but hindering ability to generalize to another task distribution. The second observation is that, by leveraging the statistics of the test data, transductive fine-tuning continues to achieve higher results than the standard fine-tuning and meta-learning, as previously reported \cite{dhillonfsbaseline}. While transductive fine-tuning, however, assumes that all the queries are available as unlabeled data. The third observation is that the accuracy of most methods on the benchmark continues to be dependent on how similar the dataset is to ImageNet: \textit{CropDiseases} commands the highest performance on average, while \textit{EuroSAT} follows in 2$^{nd}$ place, \textit{ISIC} in 3$^{rd}$, and \textit{ChestX} in 4$^{th}$. This further supports the motivation behind benchmark design in targeting applications with increasing visual domain dissimilarity to natural images. 

Table \ref{tab:classifier} shows results from varying the classifier. While mean-centriod classifier and cosine-similarity classifier are shown to be more efficient than simple linear classifier in the conventional few-shot learning setting, our results show that mean-centroid and cosine-similarity classifier only have a marginal advantage on \textit{ChestX} and \textit{EuroSAT} over linear classifier in the 5-shot case (Table \ref{tab:classifier}). As the shot increases, linear classifier begins to dominate mean-centroid and cosine-similarity classifier. One plausible reason is that since both mean-centroid and cosine-similarity classifier conduct classification based on unimodal class prototypes, when the number of examples increases, unimodal distribution becomes less suitable, and multi-modal distribution is required.

\begin{table*}[t]
\centering
\scalebox{0.7}{
\begin{tabular}{c c c c | c c c}
    \hline
    \hline
\textbf{Methods} &\multicolumn{3}{c}{\textbf{ChestX}}& \multicolumn{3}{c}{\textbf{ISIC}}  \\
    \cline{2-7}
    & 5-way 5-shot &5-way 20-shot&5-way 50-shot&5-way 5-shot &5-way 20-shot &5-way 50-shot \\
    \cline{2-7}
\emph{Linear} & 25.97\% $\pm$ 0.41\%& 31.32\% $\pm$ 0.45\%& \textbf{35.49\% $\pm$ 0.45\%}&\textbf{48.11\% $\pm$ 0.64\% }&\textbf{59.31\% $\pm$ 0.48\%}&\textbf{66.48\% $\pm$ 0.56\%}\\

\emph{Mean-centroid} &  26.31\% $\pm$ 0.42\%&30.41\% $\pm$ 0.46\%& 34.68\% $\pm$ 0.46\%& 47.16\% $\pm$ 0.54\%&56.40\% $\pm$ 0.53\% & 61.57\% $\pm$ 0.66\% \\
  
\emph{Cosine-similarity} &\textbf{26.95\% $\pm$ 0.44\%} &\textbf{32.07\% $\pm$ 0.55\%} & 34.76\% $\pm$ 0.55\%& 48.01\% $\pm$ 0.49\% &58.13\% $\pm$ 0.48\%&62.03\% $\pm$ 0.52\%\\

    \hline
\end{tabular}}
\end{table*}

\begin{table*}[t]
\centering
\scalebox{0.7}{
\begin{tabular}{c c c c | c c c}
    \hline
    \hline
\textbf{Methods} &\multicolumn{3}{c}{\textbf{EuroSAT}}& \multicolumn{3}{c}{\textbf{CropDiseases}}  \\
    \cline{2-7}
    & 5-way 5-shot &5-way 20-shot&5-way 50-shot&5-way 5-shot &5-way 20-shot &5-way 50-shot \\
    \cline{2-7}
\emph{Linear} &79.08\% $\pm$ 0.61\% &\textbf{87.64\% $\pm$ 0.47\%}& \textbf{91.34\% $\pm$ 0.37\%}&\textbf{89.25\% $\pm$ 0.51\%} &\textbf{95.51\% $\pm$ 0.31\%}&  \textbf{97.68\% $\pm$ 0.21\%}\\

\emph{Mean-centroid} &\textbf{82.21\% $\pm$ 0.49\%} &87.62\% $\pm$ 0.34\%&88.24\% $\pm$ 0.29\% & 87.61\% $\pm$ 0.47\%&93.87\% $\pm$ 0.68\%&94.77\% $\pm$ 0.34\%\\
  
\emph{Cosine-similarity} &81.37\% $\pm$ 1.54\% &86.83\% $\pm$ 0.43\%  & 88.83\% $\pm$ 0.38\% &89.15\% $\pm$ 0.51\% &93.96\% $\pm$ 0.46\%&94.27\% $\pm$ 0.41\%\\
    \hline
\end{tabular}}
\caption{The results of varying the classifier for fine-tuning on the proposed benchmark.}
\label{tab:classifier}
\end{table*}

\begin{figure*}[!th]
    \centering
    \includegraphics[width=0.3\textwidth]{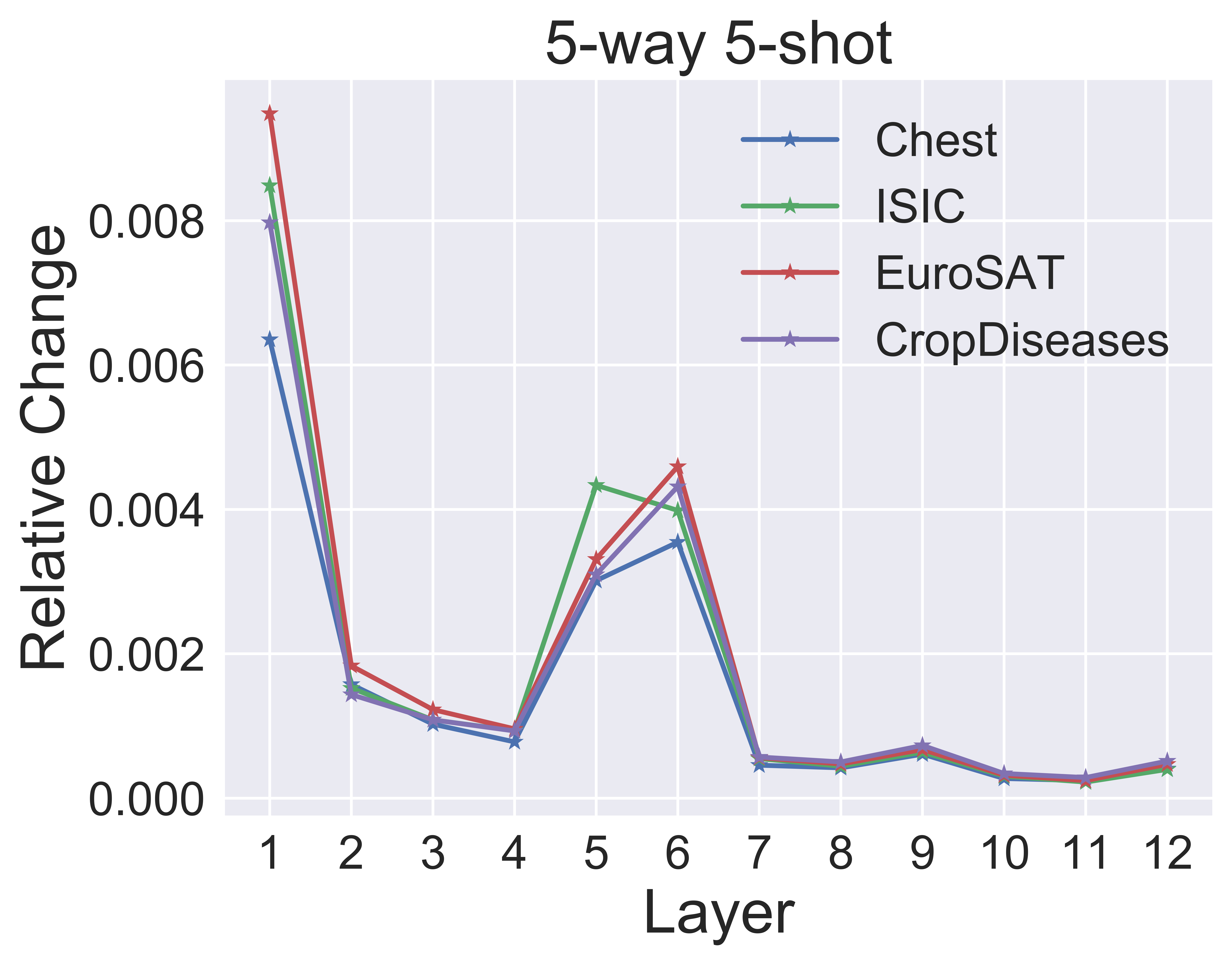}
    \includegraphics[width=0.3\textwidth]{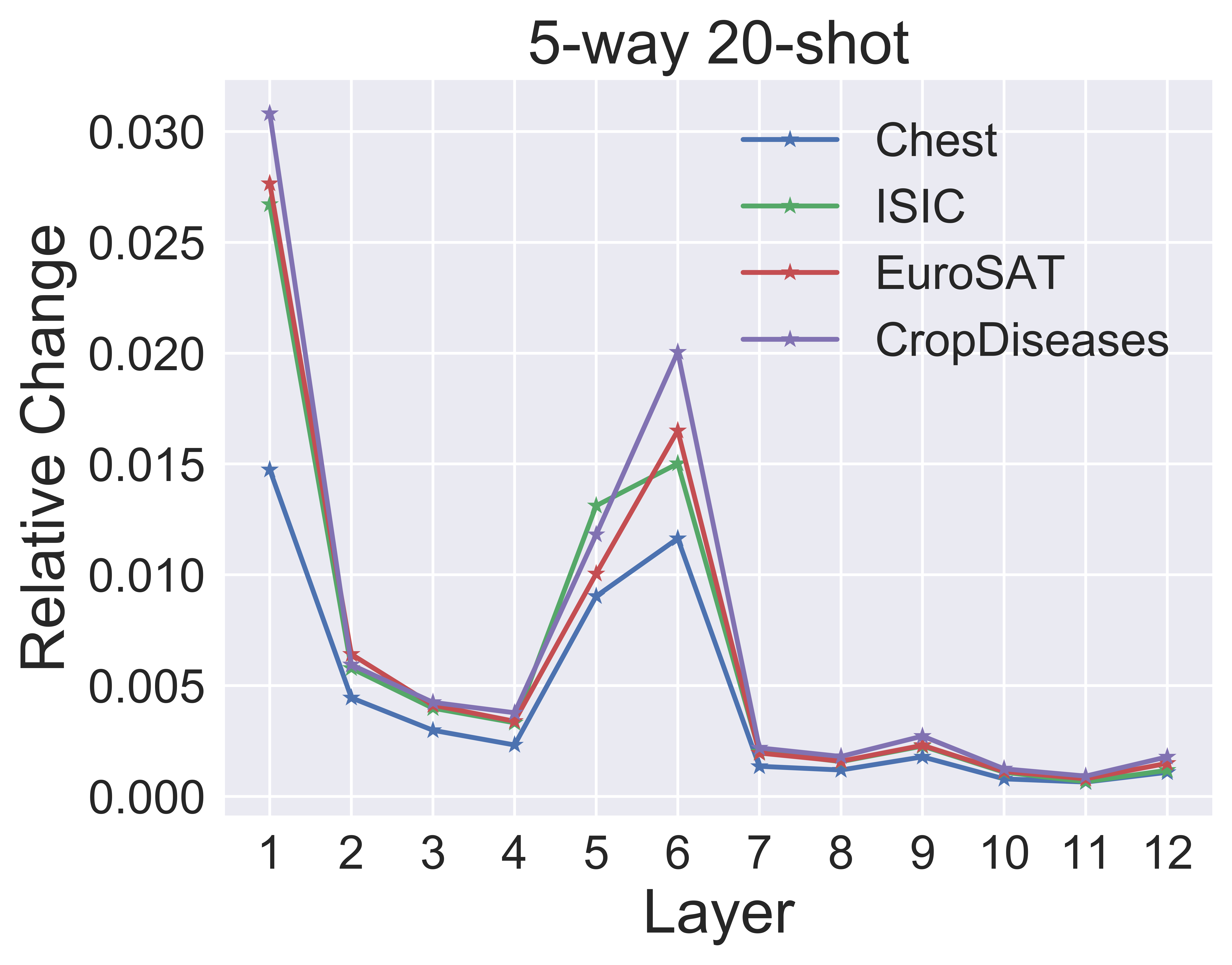}
        \includegraphics[width=0.3\textwidth]{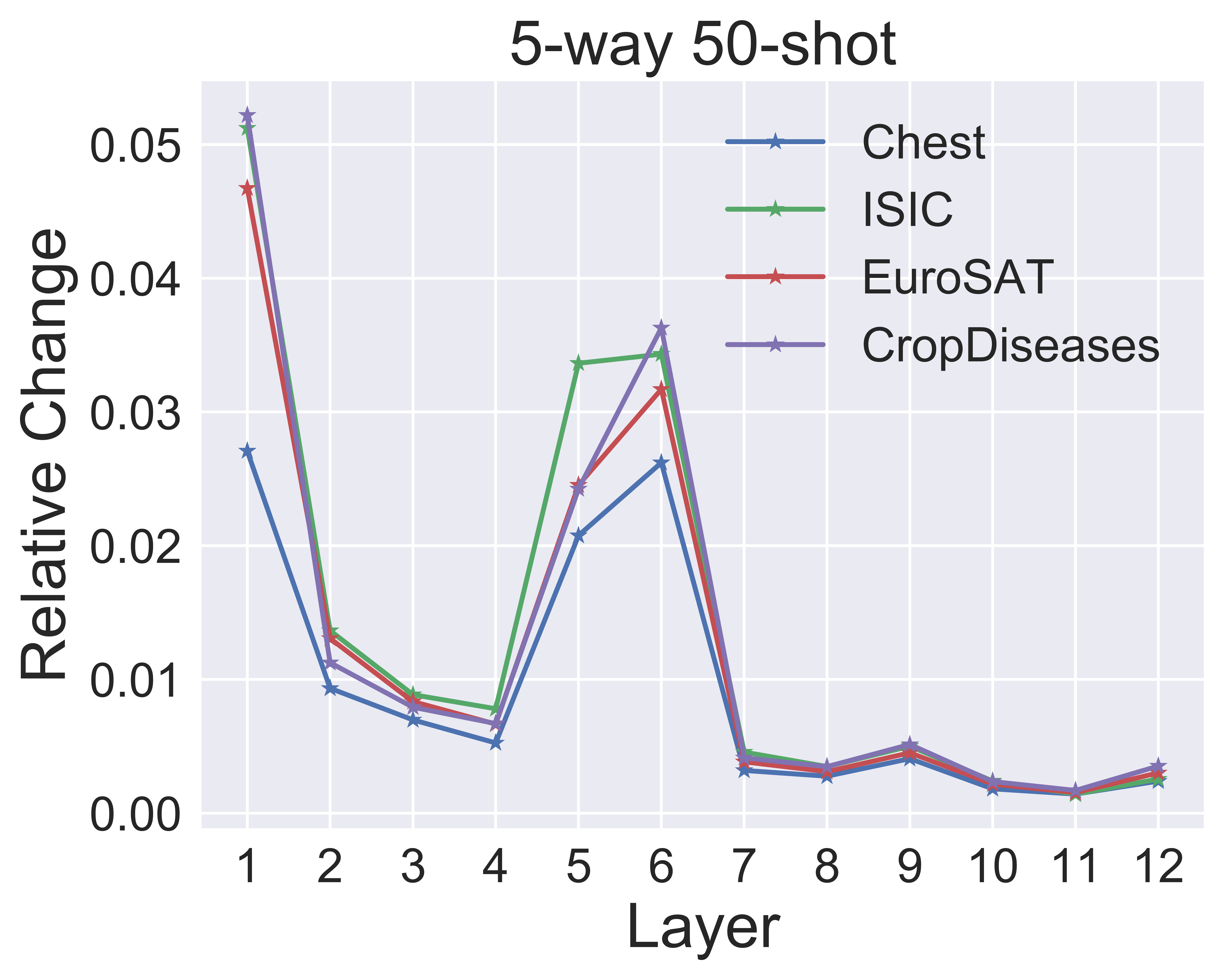}
    \caption{Relative change of pre-trained network layers for single model transfer.}
    \label{fig:layer_transfer}
\end{figure*}

We further analyze how layers are changed during transfer. We use $\theta$ to denote the original pre-trained parameters and $\hat{\theta}$ to denote the parameters after fine-tuning. Figure \ref{fig:layer_transfer} shows the relative parameter change of the ResNet10 miniImageNet pre-trained model as $\frac{|\theta -\hat{\theta}|}{|\theta|}$, averaged over all parameters per layer, and 100 runs. Several interesting observations can be made from these results. First, across all the datasets and all the shots, the first layer of the pre-trained model changes most. This indicates that if the target domain is different from the source domain, the lower layers of the pre-trained models still need to be adjusted. Second, while the datasets are drastically different, we observe that some layers are consistently more \textit{transferable} than other layers. One plausible explanation for this phenomenon is the heterogeneous characteristic of layers in overparameterized deep neural networks \cite{zhang2019all}.

\begin{table*}[t]
\centering
\scalebox{0.7}{
\begin{tabular}{c c c c | c c c}
    \hline
    \hline
\textbf{Methods} &\multicolumn{3}{c}{\textbf{ChestX}}& \multicolumn{3}{c}{\textbf{ISIC}}  \\
    \cline{2-7}
    & 5-way 5-shot &5-way 20-shot&5-way 50-shot&5-way 5-shot &5-way 20-shot &5-way 50-shot \\
    \cline{2-7}
\emph{All embeddings}  & \textbf{26.74\% $\pm$ 0.42\% }&\textbf{32.77\% $\pm$ 0.47\%}&\textbf{38.07\% $\pm$ 0.50\%}  & \textbf{46.86\% $\pm$ 0.60\%}&58.57\% $\pm$ 0.59\%& 66.04\% $\pm$ 0.56\% \\
  
\emph{IMS-f}  &25.50\% $\pm$ 0.45\% &31.49\% $\pm$ 0.47\%& 36.40\% $\pm$ 0.50\% & 45.84\% $\pm$ 0.62\% &\textbf{61.50\% $\pm$ 0.58\%}& \textbf{68.64\% $\pm$ 0.53\%}\\

    \hline
\end{tabular}}
\end{table*}
\begin{table*}[t]
\centering
\scalebox{0.7}{
\begin{tabular}{c c c c | c c c}
    \hline
    \hline
\textbf{Methods} &\multicolumn{3}{c}{\textbf{EuroSAT}}& \multicolumn{3}{c}{\textbf{CropDiseases}}  \\
    \cline{2-7}
    & 5-way 5-shot &5-way 20-shot&5-way 50-shot&5-way 5-shot &5-way 20-shot &5-way 50-shot \\
    \cline{2-7}
\emph{All embeddings}  &  81.29\% $\pm$ 0.62\%&89.90\% $\pm$ 0.41\%&92.76\% $\pm$ 0.34\% &  \textbf{90.82\% $\pm$ 0.48\% }& 96.64\% $\pm$ 0.25\%&98.14\% $\pm$ 0.18\%\\
  
\emph{IMS-f} & \textbf{83.56\% $\pm$ 0.59\%} & \textbf{91.22\% $\pm$ 0.38\%}& \textbf{93.85\% $\pm$ 0.30\%}& 90.66\% $\pm$ 0.48\%& \textbf{97.18\% $\pm$ 0.24\% }&\textbf{ 98.43\% $\pm$ 0.16\%}\\

    \hline
\end{tabular}}
\caption{The results of using all embeddings, and the  \textit{Incremental Multi-model Selection} (IMS-f) based on fine-tuned pre-trained models on the proposed benchmark.}
\label{tab:ims}
\end{table*}

% the proposed \textit{Incremental Multi-model Selection} (IMS) 
\subsubsection{Transfer from Multiple Pre-trained Models}
\label{subsec:multi}
% We include two variants of the proposed method: \emph{IMS} and \emph{IMS-f}. \emph{IMS} leverages the pre-trained models without fine-tuning on the support set while
% 67.55\% (0.45) for \emph{IMS},

The results of the described \textit{Incremental Muiti-model Selection} are shown in Table \ref{tab:ims}. \emph{IMS-f} fine-tunes each pre-trained model before applying the model selection. We include a baseline called \emph{all embeddings} which concatenates the feature vectors generated by all the layers from the fine-tuned models. Across all datasets and shot levels, the average accuracies (and 95\% confidence internals) are 68.22\% (0.45) for \emph{all embeddings}, and 68.69\% (0.44) for \emph{IMS-f}. The results show that \emph{IMS-f} generally improves upon \emph{all embeddings} which indicates the importance of selecting relevant pre-trained models to the target dataset. Model complexity also tends to decrease by over 20\% compared to \emph{all embeddings} on average. We can also observe that it is beneficial to use multiple pre-trained models than using just one model, even though these models are trained from data in different domains and different image types. Compared with standard finetuning with a linear classifier, the average improvement of \emph{IMS-f} across all the shots on \textit{ChestX} is 0.20\%, on \textit{ISIC} is 0.69\%, on \textit{EuroSAT} is 3.52\% and on \textit{CropDiseases} is 1.27\%.

In further analysis, we study the effect of the number of pre-trained models for the studied multi-model selection method. We consider libraries consisting of two, three, four, and all five pre-trained models. The pre-trained models are added into the library in the order of \textit{ImageNet}, \textit{CIFAR100}, \textit{DTD}, \textit{CUB}, \textit{Caltech256}. For each dataset, the experiment is conducted on 5-way 50-shot with 600 episodes. The results are shown in Table \ref{tab:num}. As more pre-trained models are added into the library, we can observe that the test accuracy on \textit{ChestX} and \textit{ISIC} gradually improves which can be attributed to the diverse features provided by different pre-trained models. However, on \textit{EuroSAT} and \textit{CropDiseases}, only a marginal improvement can be observed. One possible reason is that the features from \textit{ImageNet} already captures the characteristics of the datasets and more pre-trained models does not provide additional information.

\begin{figure*}[!th]
    \centering
    \includegraphics[width=0.24\textwidth]{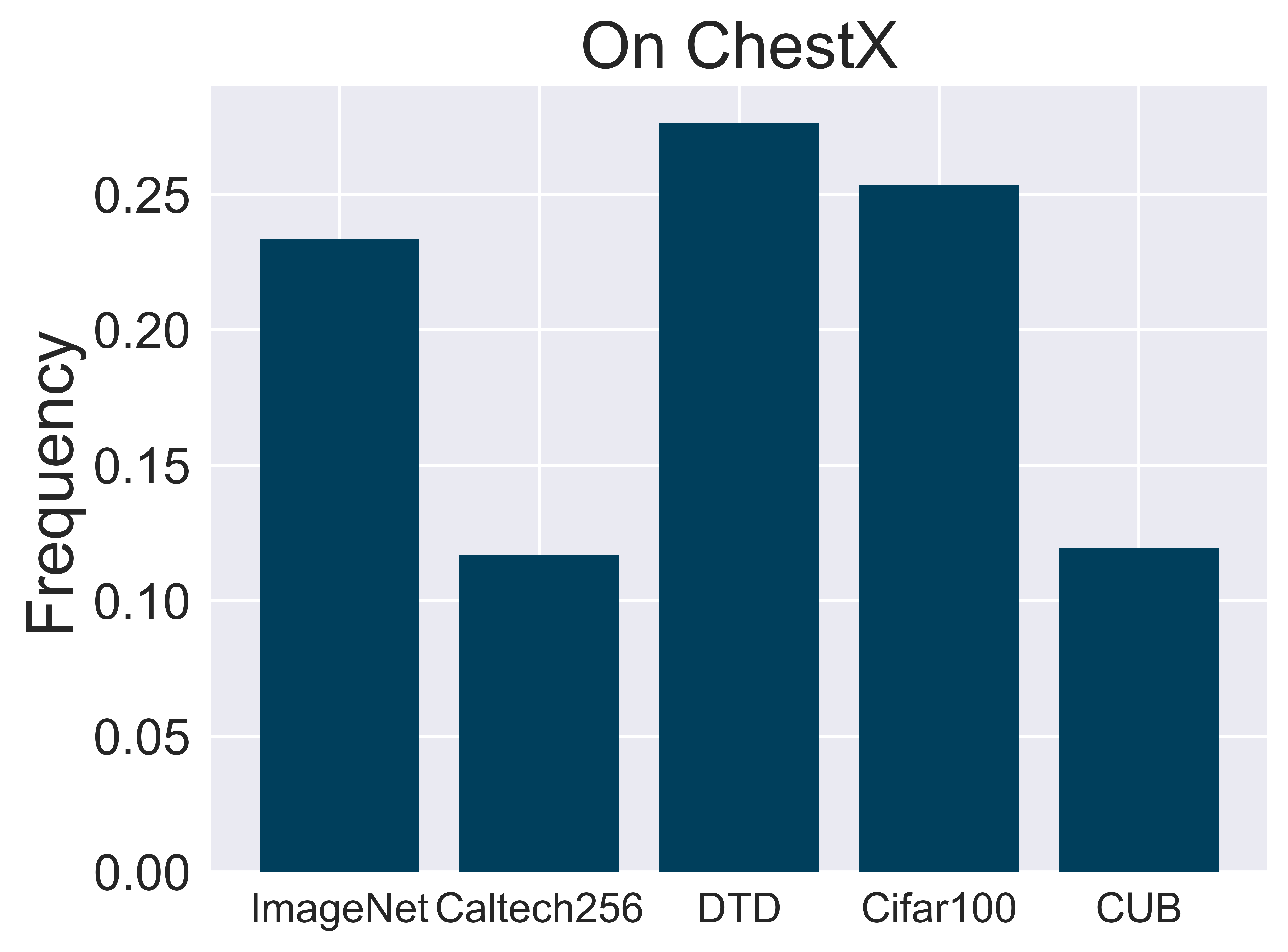}
        \includegraphics[width=0.24\textwidth]{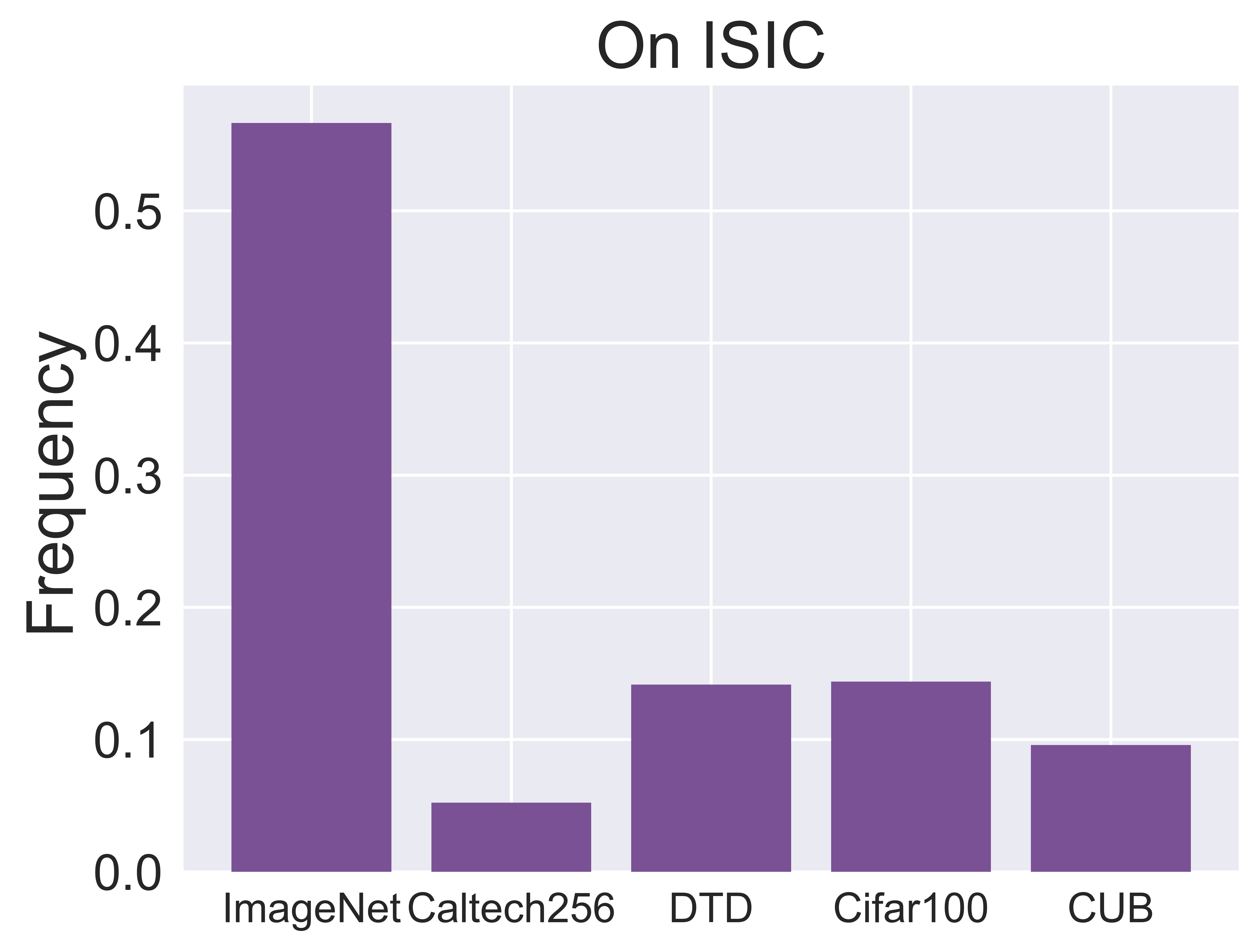}
    \includegraphics[width=0.24\textwidth]{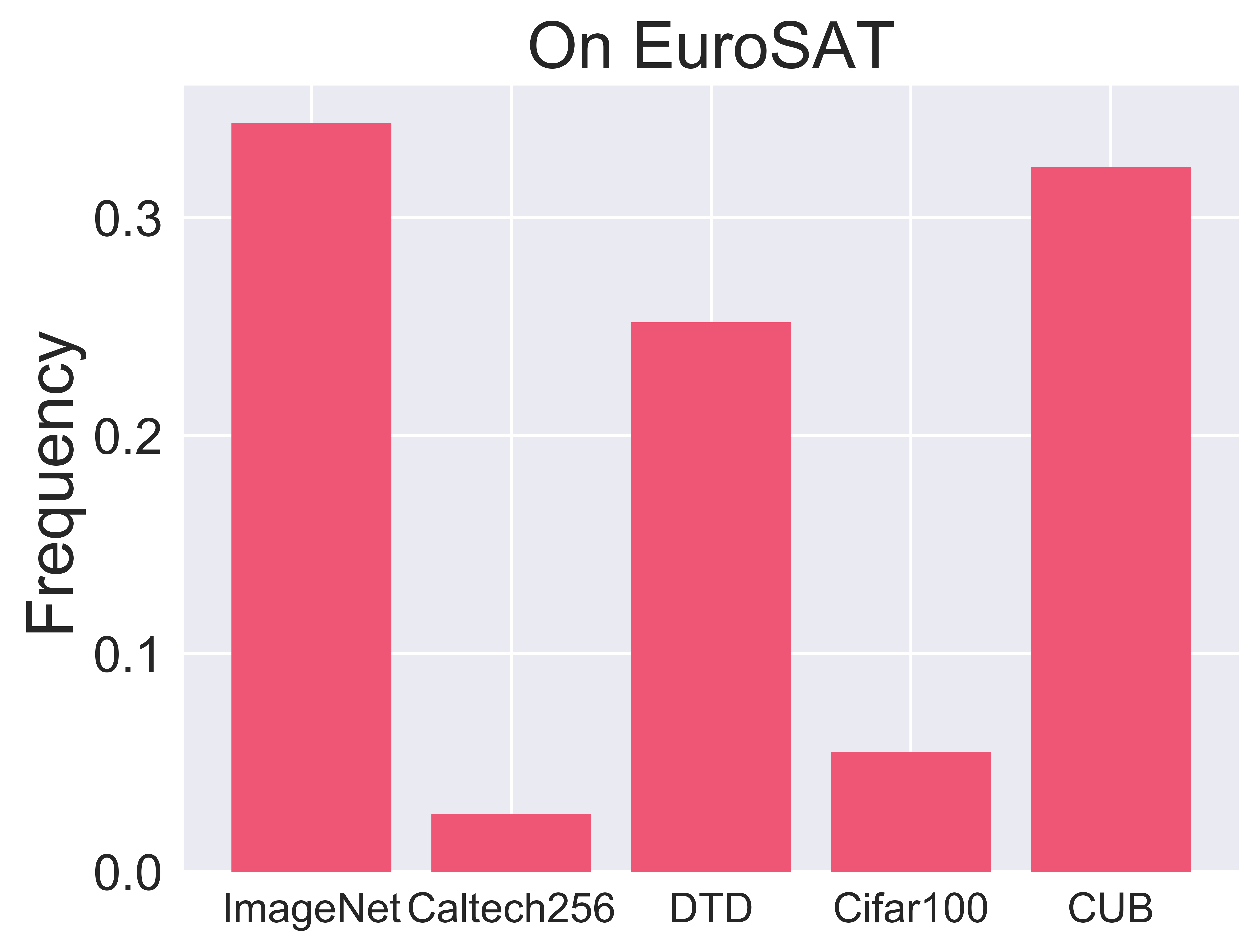}
        \includegraphics[width=0.24\textwidth]{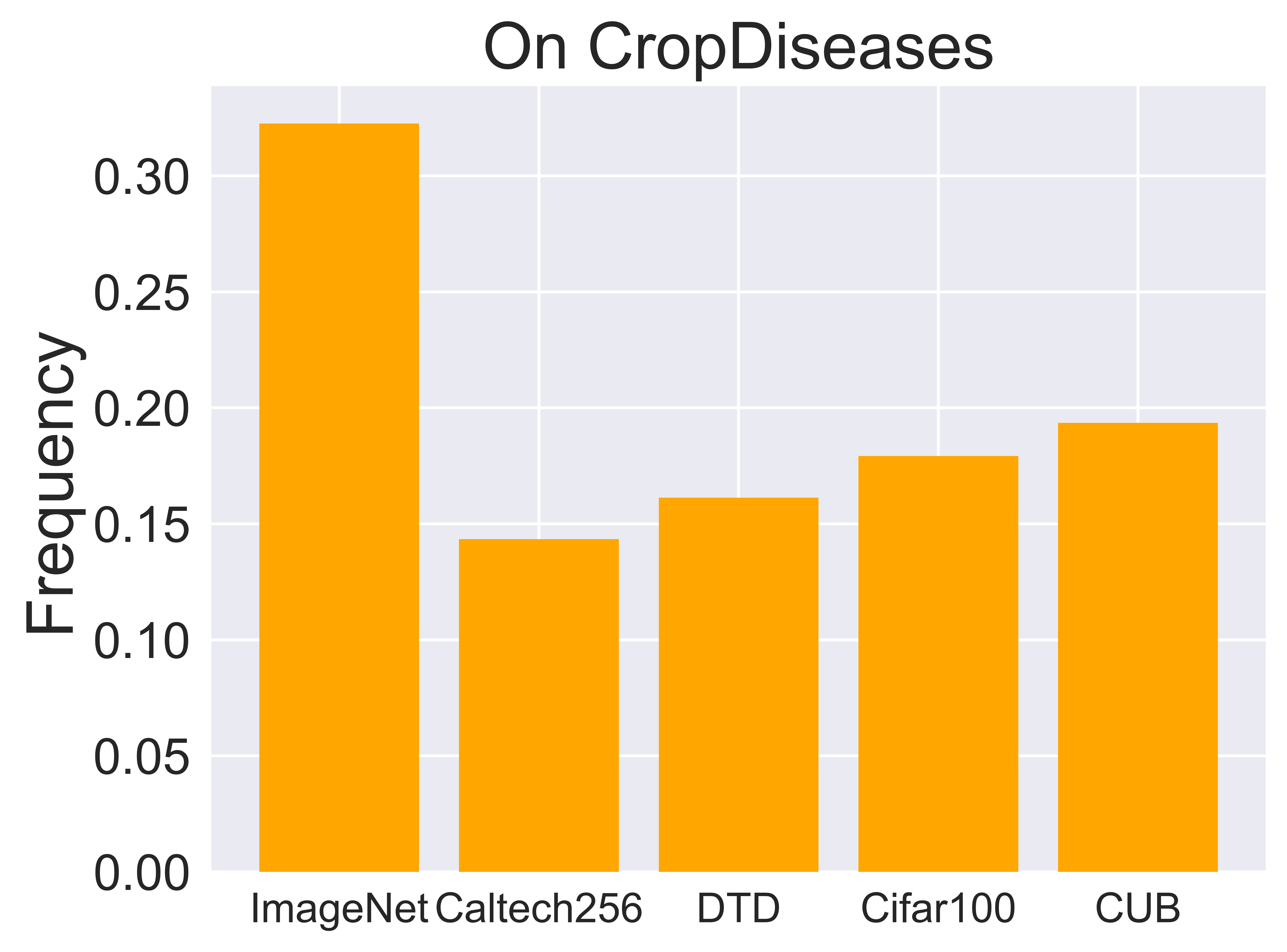}
    \caption{Frequency of source model selection for each dataset in the benchmark.}
    \label{fig:selection}
\end{figure*}

Finally, we visualize for each dataset which pre-trained models are selected in the studied incremental multi-model selection. The experiments are conducted on 5-way 50-shot with all five pre-trained models. For each dataset, we repeat the experiments for 600 episodes and calculate the frequency of each model being selected. The results are shown in Figure \ref{fig:selection}. We observe the distribution of the frequency differs significantly across datasets, as target datasets can benefit from different pre-trained models. 

\iffalse
\begin{table}[!th]
\centering
\scalebox{0.8}{
\begin{tabular}{c |c | c | c| c }
    \hline
    \hline
    \diagbox{Dataset}{\emph{\# of models}}
  &  2 & 3 & 4 & 5
\\
    \hline
\textbf{ChestX}  &  34.35\%    & 36.29\%   & 37.64\%   &  37.89\%  
\\
    \hline
\textbf{ISIC}  & 59.4\%    & 62.49\%   & 65.07\%    &  64.77\%  
\\
    \hline
\textbf{EuroSAT}  &  91.71\%   & 93.49\%   & 92.67\%   & 93.00\%  
\\
    \hline
\textbf{CropDiseases}  & 98.43\%    & 98.09\%    & 98.05\%   & 98.60\%  
\\
    \hline
\end{tabular}}
\caption{Number of models' effect on test accuracy.}
\label{tab:num}
\end{table}

\begin{figure}[!th]
    \centering
    \includegraphics[width=0.5\linewidth]{./figs/conclusion.png}
    \caption{Best meta-learning, single model, and multi-model transfer learning.}
    \label{fig:inter}
\end{figure}
\fi

\begin{table}[!t]

	\begin{minipage}{0.45\linewidth}

    \scalebox{0.7}{\begin{tabular}{c |c | c | c| c }
    \hline
    \hline
    \diagbox{Dataset}{\emph{\# of models}}
  &  2 & 3 & 4 & 5
\\
    \hline
\textbf{ChestX}  &  34.35\%    & 36.29\%   & 37.64\%   &  37.89\%  
\\
    \hline
\textbf{ISIC}  & 59.4\%    & 62.49\%   & 65.07\%    &  64.77\%  
\\
    \hline
\textbf{EuroSAT}  &  91.71\%   & 93.49\%   & 92.67\%   & 93.00\%  
\\
    \hline
\textbf{CropDiseases}  & 98.43\%    & 98.09\%    & 98.05\%   & 98.60\%  
\\
    \hline
		\end{tabular}}
		\caption{Number of models' effect on test accuracy.}
			\label{tab:num}
	\end{minipage}\hfill
	\begin{minipage}{0.45\linewidth}
\includegraphics[width=1.0\linewidth]{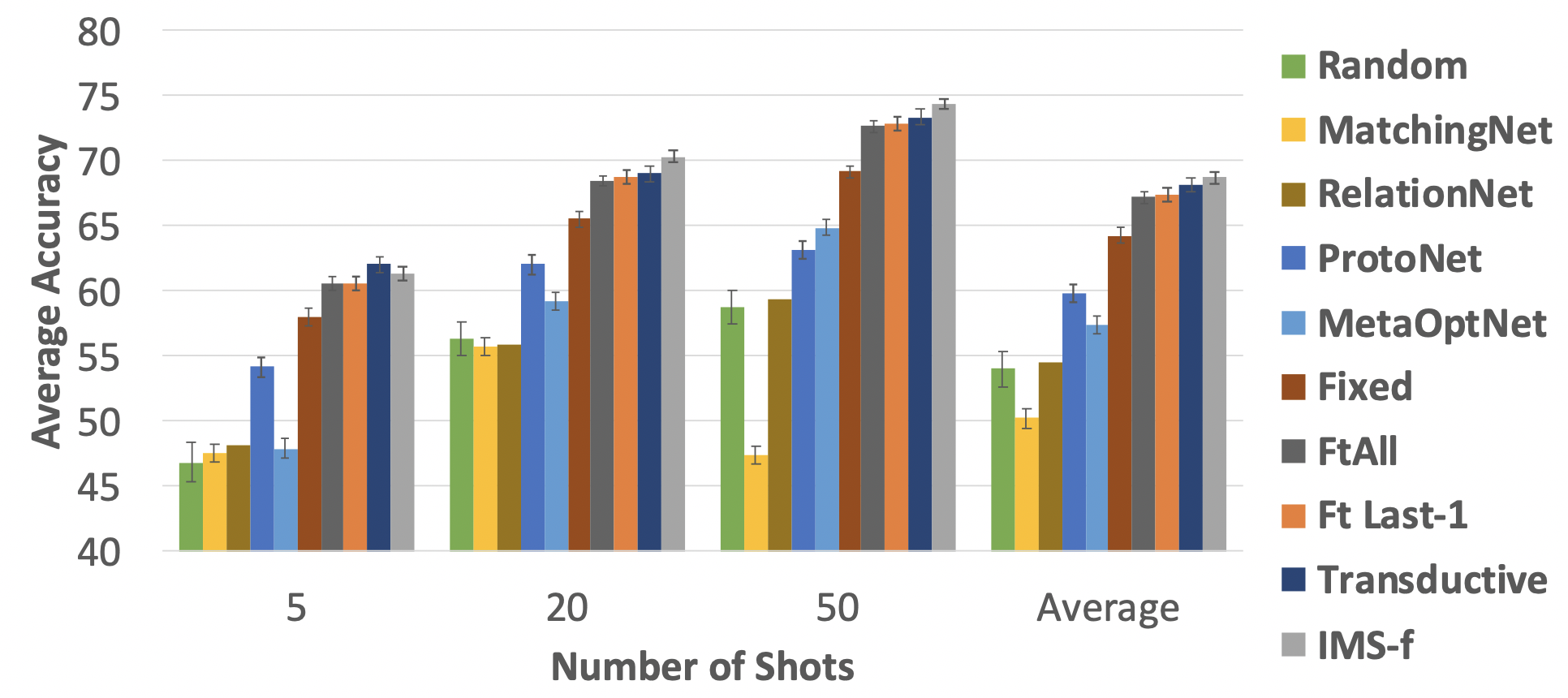}
    \captionof{figure}{Comparisons of methods across the entire benchmark.}
    \label{fig:inter}
	\end{minipage}
\end{table}

\subsection{Benchmark Summary}
\label{subsec:inter}

Figure \ref{fig:inter} summarizes the comparison across algorithms, according to the average accuracy across all datasets and shot levels in the benchmark. The degradation in performance suffered by meta-learning approaches is significant. In some cases, a network with random weights outperforms meta-learning approaches. FWT methods, which yielded no performance improvements, are omitted for brevity. MAML, which failed to operate on the entire benchmark, is also omitted.
%In addition, the relative performance gain with increasing number of shots is greater with transfer methods compared to meta-learning: 5.8\% and 5.7\% from 20-shot to 50-shot for single and multi-model transfer learning, respectively, versus 1.8\% for meta-learning. 5-shot to 20-shot was similar for all methods: 14.5\%, 13.6\%, 14.6\%, for meta-learning, single model, and multi-model, respectively.

\section{Conclusion}
In this paper, we formally introduce the Broader Study of Cross-Domain Few-Shot Learning (BSCD-FSL) benchmark, which covers several target domains with varying similarity to natural images. We extensively analyze and evaluate existing meta-learning methods, including approaches specifically designed for cross-domain few-shot learning, and variants of transfer learning. The results show that, surprisingly, state-of-art meta-learning approaches are outperformed by earlier approaches, and recent methods for cross-domain few-shot learning actually degrade performance. In addition, all meta-learning methods significantly underperform in comparison to fine-tuning methods. In fact, some meta-learning approaches are outperformed by networks with random weights. In addition, accuracy of all methods correlate with proposed measure of data similarity to natural images, verifying the diversity of the proposed benchmark in terms of its problem representation, and its value towards guiding future research. In conclusion, we believe this work will help the community understand what methods are most effective in practice, and help drive further advances that can more quickly yield benefit for real-world applications. 

\section{Acknowledgement}
This material is based upon work supported by the Defense Advanced Research Projects Agency (DARPA) under Contract No. FA8750-19-C-1001. Any opinions, findings and conclusions or recommendations expressed in this material are those of the author(s) and do not necessarily reflect the views of the Defense Advanced Research Projects Agency (DARPA). This work was supported in part by CRISP, one of six centers in JUMP, an SRC program sponsored by DARPA.  This work is also supported by NSF CHASE-CI \#1730158, NSF FET \#1911095, NSF CC* NPEO \#1826967.

%\iffalse
\newpage
\bibliographystyle{splncs04}
\bibliography{ref}

\newpage
\section{Appendix}

\subsection{Incremental Multi-model Selection}

\begin{algorithm}[h]
\DontPrintSemicolon

  \tcc{First stage}
  $I_1$ = \{\}
  
    \tcc{Iterate over each pre-trained model}
    
    \For{$c = 1$ $\rightarrow$ $C$}    
    { 
        $min\_loss$ = -1
        
        $best\_l$ = $None$
        
        \tcc{Iterate over each layer of the pre-trained model}
        
        \For{$l = 1$ $\rightarrow$ $L$}    
        {
            \If{$CW(S, \{l\})$ $<$ $min\_loss$ }
            {  
                $best\_l$ = $l$
                
                $min\_loss$ = $CW(S, \{l\})$
            }
        }
        $I_l$ = $I_l$ $\bigcup$ $best\_l$
    }
\tcc{Second stage}
    $I$ = \{\}
    
    $min\_loss$ = -1
    
    \For{\textnormal{each} $l$ \textnormal{in} $I_1$}    
        {
            \If{$CW(S, I \bigcup l)$ $<$ $min\_loss$ }
            {  
                $min\_loss$ = $CW(S, I \bigcup l)$
                 
                $I$ = $I$ $\bigcup$ $l$
            }
        }
        
    Concatenate the feature vectors generated by the layers in $I$ and train a linear classifier.

\caption{\emph{Incremental Multi-model Selection}. $S = \{x_i, y_i\}_{i=1}^{K \times N}$ is a support set consisting of $N$ examples from $K$ novel classes. Assume there is a library of $C$ pre-trained models $\{M_c\}^C_{c=1}$. Each model has $L$ layers and $l$ is used to denote one particular layer. Let $CW(S, I)$ be a function which returns the average cross-validation error given a dataset $S$ and a set of layers $I$ which are used to generate feature vector.} 
\label{alg: ims}
\end{algorithm}
\end{document}